\documentclass[10pt,twocolumn,letterpaper]{article}

\usepackage[pagenumbers]{cvpr} 

\usepackage{colortbl}
\usepackage{color}
\usepackage{CJKutf8}
\definecolor{cvprblue}{rgb}{0.21,0.49,0.74}
\usepackage[pagebackref,breaklinks,colorlinks,allcolors=cvprblue]{hyperref}
\usepackage{multirow} 
\def\paperID{1875} 
\def\confName{CVPR}
\def\confYear{2025}
\title{Cross-modal Causal Relation Alignment for Video Question Grounding}

\author{
Weixing Chen$^1$ \hspace{0.7em} Yang Liu$^1$\thanks{Corresponding Author} \hspace{0.7em}  Binglin Chen$^1$ \hspace{0.7em}  Jiandong Su$^2$ \hspace{0.7em}  Yongsen Zheng$^3$ \hspace{0.7em}  Liang Lin$^1$\\
$^1$Sun Yat-sen University $^2$Shenzhen Institute of Advanced Technology \\$^3$Nanyang Technological University, Singapore\\
{\tt\small \{chenwx228,chenblin9\}@mail2.sysu.edu.cn,liuy856@mail.sysu.edu.cn}\\
{\tt\small jiandong.laurence.su@gmail.com,yongsen.zheng@ntu.edu.sg,linliang@ieee.org}
}
\begin{document}
\maketitle

\begin{abstract}
Video question grounding (VideoQG) requires models to answer the questions and simultaneously infer the relevant video segments to support the answers. However, existing VideoQG methods usually suffer from spurious cross-modal correlations, leading to a failure to identify the dominant visual scenes that align with the intended question. Moreover, vision-language models exhibit {\color{black}unfaithful} generalization performance and lack robustness on challenging downstream tasks such as VideoQG. In this work, we propose a novel VideoQG framework named Cross-modal \textbf{C}ausal \textbf{R}elation \textbf{A}lignment (\textbf{CRA}), to eliminate spurious correlations and improve the causal consistency between question-answering and video temporal grounding. Our CRA involves three essential components: i) Gaussian Smoothing Grounding (GSG) module for estimating the time interval via cross-modal attention, which is de-noised by an adaptive Gaussian filter, ii) Cross-Modal Alignment (CMA) enhances the performance of weakly supervised VideoQG by leveraging bidirectional contrastive learning between estimated video segments and QA features, iii) Explicit Causal Intervention (ECI) module for multimodal deconfounding, which involves front-door intervention for vision and back-door intervention for language. Extensive experiments on two VideoQG datasets demonstrate the superiority of our CRA in discovering visually grounded content and achieving robust question reasoning. Codes are available at \url{https://github.com/WissingChen/CRA-GQA}.
\end{abstract}

\section{Introduction}

Recent advancements in vision-language models (VLMs) have significantly improved the performance of Video Question Answering (VideoQA) tasks~\cite{yu2024self, min2024morevqa}. However, these improvements are not always faithful, as models may rely on statistical biases (i.e., language/visual short-cut or spurious vision-language correlation) in the training data rather than genuine causal visual evidence (i.e., dominant visual content relevant to the prediction)~\cite{xiao2024can}. {\color{black} To ensure faithful performance enhancement, the Video Question Grounding (VideoQG) task has emerged to explicitly provide visual evidence for the given answers~\cite{qian2023locate} by highlighting the relevance between the visual grounding content and the question reasoning process.}


\begin{figure}[!t]
    \centering
    \includegraphics[width=1\linewidth]{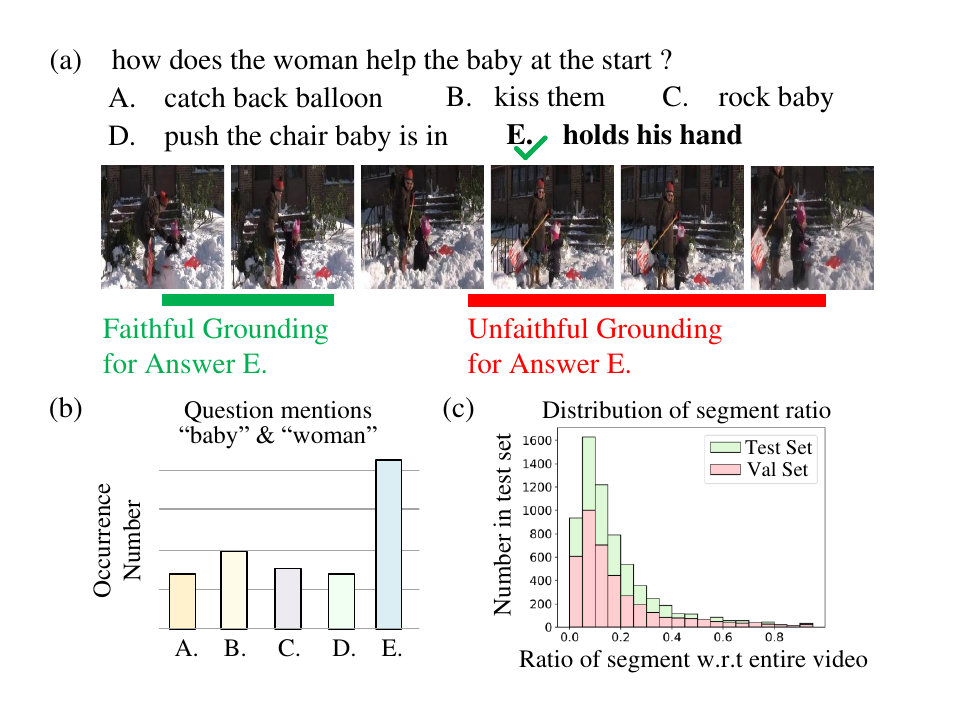}
    \caption{(a) A typical example of a VideoQG task, which adopts the erroneous grounding and leads to the correct but unfaithful answer. (b) shows the occurrence number of different answers from the questions that mention ``baby" and ``woman". (c) shows the distribution of the ratio between the video segment and full video in the Test set and Val Set.}
    \label{fig:intro_causal}
\end{figure}

{\color{black}Generally, VideoQG models need to be trained using annotated time intervals derived from the VideoQA task~\cite{xu2022hisa}. Nevertheless, the adopted datasets often lack annotations related to video grounding due to the high labeling costs ~\cite{xiao2024can}.}
Although current video language models can {\color{black}also} perform zero-shot VideoQG~\cite{zhang2023simple, wang2024hawkeye, fei2024video}, these capabilities largely rely on extensive training with pre-aligned, curated video-text datasets. This dependence suggests that such models may be less effective for the out-of-domain (OOD) downstream tasks like VideoQG ~\cite{varma2024ravl}. {\color{black} Additionally}, the inherent hallucination tendencies of {\color{black}these large video language models} can introduce significant noise, potentially compromising model reliability~\cite{patraucean2024perception}.
Given the abundance of existing VideoQA datasets~\cite{jang2017tgif, yu2019activitynet, xiao2021next}, it is essential for a VideoQG model that could leverage only visual features of videos and textual features of QA pairs in VideoQA datasets to achieve VideoQG. Specifically, the VideoQG model can estimate video segments related to the QA task using multiple pre-processed candidate time intervals, known as the two-stage approach~\cite{li2022equivariant}. {\color{black} Moreover}, the one-stage VideoQG model~\cite{xiao2024can} can also directly estimate the time intervals, which involves simpler modeling and efficient reasoning.


{\color{black} However}, due to the biases present in videos and language, VideoQG models may rely on these spurious correlations caused by multi-modal biases during training. These spurious correlations make the models rely on the unfaithful visual grounded scene to answer questions rather than the causal visual grounded scene~\cite{wei2023visual, chen2023cross,liu2023cross}. As shown in Figure~\ref{fig:intro_causal} (a), the NextGQA dataset presents a video $V$ depicting an interaction between a ``\textit{baby}" and a ``\textit{woman}", based on which a question $L$ with an answer set is posed. Given this information, the VideoQG is required to concurrently output the corresponding answer $a$ and the temporal interval $t$ of the video. But the existing VideoQG model outputs the correct answer $E$ based on the unfaithful visual grounding scene.  There are two major observations: \textit{1) Linguistic Bias.} By analyzing questions involving both ``\textit{baby}" and ``\textit{woman}" and their answers in the training set, we found a significant distributional disparity, i.e., data bias, as illustrated in Figure~\ref{fig:intro_causal} (b). \textit{2) Visual Bias.} Furthermore, the ratio of video segments to full videos is approximately 0.1, as shown in Figure~\ref{fig:intro_causal} (c), which makes precise grounding highly challenging for the model. Additionally, this imbalance also leads the model to establish a spurious correlation between the confounder (``\textit{baby}" and ``\textit{woman}") and the correct answer $E$ and overlook correct visual cues when giving answers. Therefore, explicitly discovering multi-modal causal relations for VideoQG is important and challenging.


To address these challenges, we utilize the structural causal model (SCM)~\cite{pearl2016causal} for analysis and construct a causal diagram to model the VideoQG task. Different from existing causality methods~\cite{li2022invariant,wei2023visual,liu2023cross} that split and recombine video segments to improve causal consistency between video and QA features, our method aims to find the explicit visual evidence (i.e., temporal grounding) that aligns with the QA semantics in an end-to-end front-door intervention framework. 
Furthermore, since visual confounders for VideoQG are difficult to describe without relying on well-trained semantic extractors~\cite{zang2023discovering}, we apply front-door intervention for visual deconfounding.
Additionally, the existing mediators in the front-door intervention are often latent features, making describing and demonstrating their mechanisms challenging. 
To achieve faithful question-answering and visual grounding, VideoQG is explicitly incorporated into the causal chain of VideoQA, which not only eliminates the need for additional annotations but also enhances interpretability.
Since linguistic confounders can be decomposed through syntactic analysis in a structured form, we use back-door intervention for language deconfounding.

To accomplish these goals, we propose a Cross-modal Causal Relation Alignment (CRA) framework, which comprises three key components: the Gaussian Smoothing Grounding (GSG) module, the Explicit Causal Intervention (ECI) module, and the
Cross-Modal Alignment (CMA) module. 
Using the VideoQA task as a weak supervisory signal, CRA finely aligns video features with QA features and guides the GSG module in generating time intervals. ECI then uses the estimated video segment as a mediator for front-door intervention, enabling the isolation and elimination of spurious correlations caused by confounders by considering both the entire video and each potential video segment. This approach not only enhances VideoQA performance but also allows the effectiveness of the intervention to be quantified through grounding performance.
Besides, by statistically analyzing entities and their semantics relations in QA, back-door intervention can treat them as confounders, mitigating spurious correlations caused by keywords.
Our CRA not only enhances the performance of VideoQG but also provides new insights into the application of causal reasoning in multimodal tasks.  The main contributions are summarized as follows:

\begin{itemize}
    \item To concurrently capture relations between Video and QA and resist temporally irrelevant noise, we propose the Gaussian Smoothing Grounding (GSG) module, which reliably generates time intervals from cross-attention. 
    \item Given the lack of VideoQG annotations and the limitations of large models in downstream VideoQG task, we introduce a bidirectional cross-modal alignment module to effectively achieve weakly supervised VideoQG. 
    \item To achieve causal relation alignment, the video segment is explicitly incorporated into the causal chain of VideoQA as a mediator, which not only enables interpretable causal intervention but also allows the effect of the intervention to be quantified by the grounding result.
\end{itemize}

\section{Related Work}
\begin{figure*}[!t]
    \centering
    \includegraphics[width=0.95\linewidth]{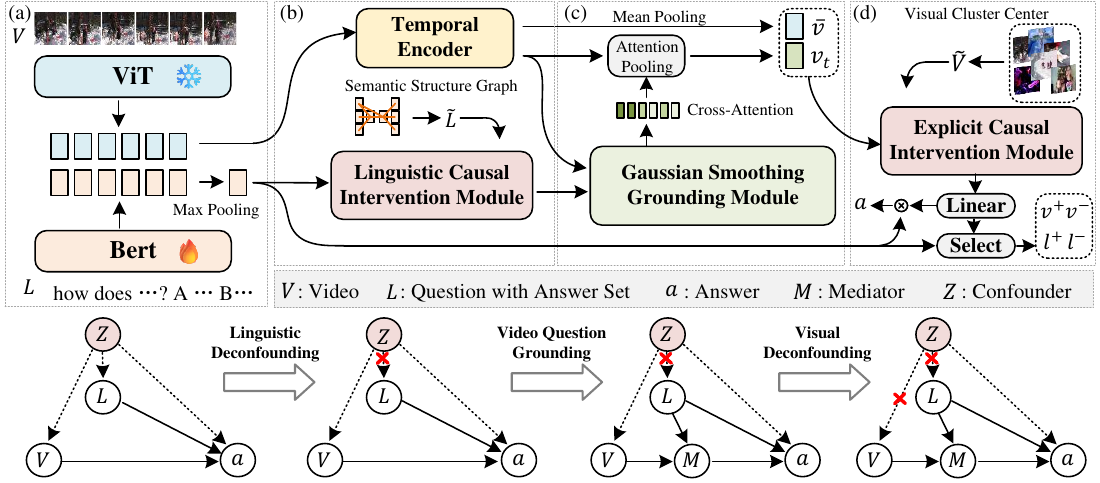}
    \caption{An overview of our CRA framework, and the above shows our proposed SCM in CRA.
    (a) It extracts video and linguistics features separately. 
    (b) A Temporal Encoder is used to fuse temporal information and the Linguistics Causal Intervention Module mitigates the bias from the QA feature using the semantic structure graphs as confounders $\widetilde{L}$. (c) Our Gaussian Smoothing Attention Grounding module estimates the cross-modal attention to refine the video feature, and then the average visual feature $\bar{V}$, grounded visual feature $M$, and the pre-processed visual feature clusters $\widetilde{V}$ are provided for the Explicit Causal Intervention Module in (d). Finally, the cross-entropy loss is computed for $a$, and bidirectional contrastive learning is applied to the selected positive and negative multi-modal samples for CMA.}
    \label{fig:method}
\end{figure*}
\subsection{Temporal Grounded Video QA}
In the Temporal Grounded Video QA task, the model must integrate video understanding, NLP, and temporal information extraction to locate the time interval in the video relevant to the question and provide accurate answers~\cite{he2019read,zhang2023temporal}.
To effectively extract fine-grained and aligned multimodal representations, numerous approaches have been proposed, including cross-modal attention~\cite{gao2019structured, li2019beyond}, memory networks~\cite{fan2019heterogeneous}, and graph reasoning~\cite{cherian20222}. 
In VideoQG, there are typically two approaches: the two-stage method~\cite{chen2020look,li2022invariant}, which involves generating candidate intervals first and then performing matching, and the one-stage method~\cite{jin2022embracing, xiao2024can}, which directly decodes the location interval from global features. 
The two-stage method generally performs better in complex scenarios, while the one-stage method, though simpler and more efficient, tends to struggle with handling long videos~\cite{pan2023scanning}. 
Current models in the Temporal Grounded Video QA task still face challenges, such as limited capability in processing long videos and complex events, as well as heavy reliance on large-scale annotated data~\cite{cao2023iterative}. 
To address these issues, we align multi-modal features in the latent space and generate temporal intervals without relying on explicit grounding annotations. We achieve temporal grounding by weak supervision from the VideoQA task and implement front-door and back-door interventions for visual and textual representations, respectively.

\subsection{Multi-modal Causality Learning}
Traditional multimodal approaches primarily rely on correlation analysis, but their neglect of causality can lead to limitations in tasks involving causal dependencies, such as the VideoQG task~\cite{liu2022causal,wei2023visual,liu2023causalvlr}. To tackle this challenge, researchers have introduced causal intervention mechanisms, including back-door intervention~\cite{wang2020visual, wang2022weakly}, front-door intervention~\cite{liu2023cross, wei2023visual}, and counterfactual reasoning~\cite{tang2023towards, zhang2023reducing}. These approaches aim to eliminate spurious correlations, thereby enhancing the model's robustness.
In the VideoQG task, causal inference aids the model in identifying causally relevant chains along the temporal dimension, enabling it to faithfully answer questions based on the video segments corresponding to these causal chains. IGV~\cite{li2022invariant} and EIGV~\cite{li2022equivariant} leverage the principle that the correlations between a causal scene and the answer should remain invariant under different complement scenes to achieve causal VideoQA. 
However, these approaches struggle to extract fine-grained conceptual semantics from the video and discover causal relations~\cite{shah2024front}. 
To overcome this, we propose bi-directional cross-modal alignment that offers a more comprehensive understanding of cross-modal relations in the video. Then, the explicit causal intervention module of the CRA framework uses the estimated video segment as a mediator front-door causal intervention that the performance of grounding can be the metric for intervention.

\section{Method}
{\color{black} To enhance the faithful performance for VideoQG}, as shown in Figure~\ref{fig:method}, our proposed CRA comprises the Gaussian Smoothing Grounding (GSG) Module, Cross-Modal Alignment (CMA) 
and the multi-modal causal intervention including Linguistic Causal Intervention (LCI) Module, and Explicit Causal Intervention (ECI) Module. The GSG module first estimates the video segment by cross-attention using the deconfounded linguistics feature and the multi-modal feature can be aligned for weakly supervised VideoQG. Finally, the ECI module applies front-door causal intervention to alleviate confounding between cross-modal features.

Given a full video $V$ and a question $L$ with a candidate answer set $A$, the VideoQG model should infer the answer $a$ and identify the time interval $t$ of the video segments that serve as the rationale for $a$. The task can be formulated as:
\begin{equation}
    a^*, t^* = \textrm{argmax}_{a\in A}\Psi(a|V, L, w)\Phi(w|V, L)
    \label{eq:VideoQG}
\end{equation}
in which $w$ is the temporal attention relevant to $L$, it can be calculated for the time interval $t$, and the grounded video feature $v_t$. $\Psi$ and $\Phi$ are modules that contain VideoQA and VideoQG counterparts of CRA, respectively.

Specifically, as shown in Figure~\ref{fig:method} (a), we first use a pre-trained CLIP model~\cite{radford2021learning} to extract features $v\in\mathbb{R}^{n\times d}$ from $n$ evenly sampled frames from the video and $d$ means the embedding size. For the language, we encode the question and answer set using the RoBERTa model~\cite{liu2019roberta} to obtain linguistics features $l\in\mathbb{R}^{m\times d}$ where $m$ is the length of the QA and then apply max pooling to achieve a global representation $l_g\in\mathbb{R}^{1\times d}$. 

\subsection{Gaussian Smoothing Grounding Module}
After feature extraction, the temporal information can be fused in $v$ using the Temporal Encoder, as shown in Figure~\ref{fig:method} (b). The Temporal Encoder consists of a 2-layer transformer, which enhances the contextual information of each frame sequence through self-attention layers. 
Then the estimation of time interval $t$ for the video segment relevant to $L$ can be implemented via cross-modal attention $w$, as shown in Figure~\ref{fig:method} (c). Similar to Temp[CLIP]~\cite{xiao2024can}, the time interval $t$ can be calculated by a post-hoc analysis of the attention from the GSG module. 

\begin{figure}[!t]
    \centering
    \includegraphics[width=1\linewidth]{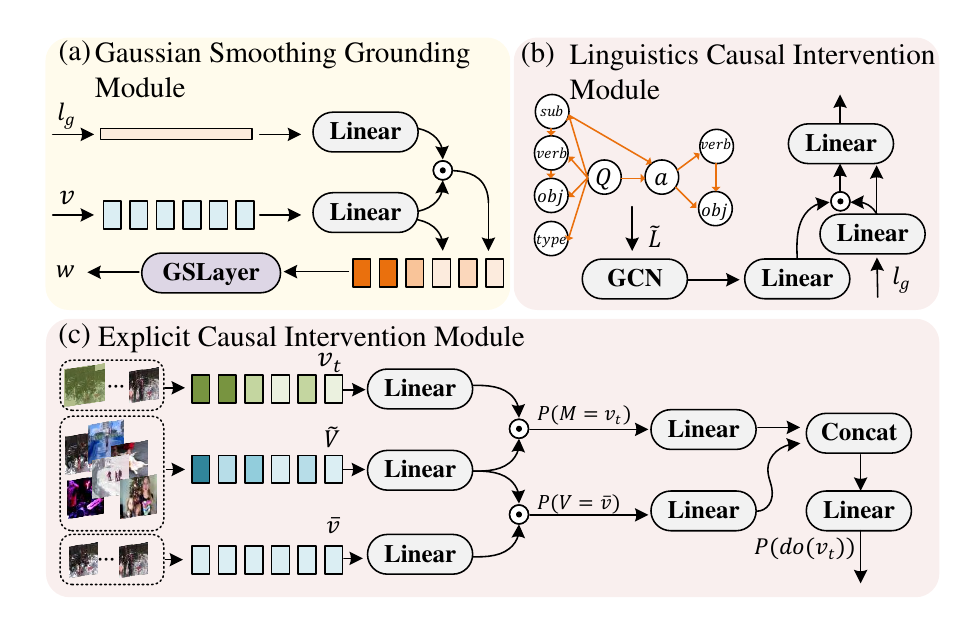}
    \vspace{-15pt}
    \caption{
    (a) The Gaussian Smoothing Grounding Module and the Multi-modal Causal Intervention Module are presented that consisting of (b) the back-door intervention module and (c) the Explicit intervention module, where $\widetilde{L}$ is the semantic graph constructed by Stanza~\cite{qi2020stanza} and $\widetilde{V}$ is constructed from all frames in the training set.
    }
    \label{fig:module}
\end{figure}

As shown in Figure~\ref{fig:module} (a), our GSG module calculates the relevance between $l_g$ and $v$ and estimates the cross-modal attention via GSLayer, which can be formulated as follows:
\begin{equation}
    w = G(\mathbf{MLP}(v \cdot l_g^T))
\end{equation}
where $G(\cdot)$ is an adaptive Gaussian filter with the learnable parameter, which is resistant to temporal attention instability.
To obtain the time interval, $w$ is summarized using attention pooling to aggregate video segment features $v_t = w\times v$ and optimized through the VideoQA task. Finally, the segment or frame with the highest attention value is identified, and thresholding around it is applied to determine $t$.


\subsection{Cross-Modal Alignment}
However, relying solely on the cross-entropy loss from the VideoQA task is insufficient to provide strong guidance for VideoQG. Therefore, we propose a bi-directional alignment method to guide our CRA in estimate the cross-modal attention.
In the VideoQG task, the grounded video features $v_t$ and language features $l_g$ are estimated, with these features then randomly sampled to produce multi-modal positive and negative samples. First, for aligning language features with visual features, following the approach by Xiao et al.~\cite{xiao2024can}, $k_l$ QA pairs from different videos within the same batch are sampled as negative samples $l^-$, while the QA pair corresponding to the current video serves as the positive sample $l^+$. Additionally, $k_v$ different grounded video features within the same batch are sampled as negative samples $v^-$, while the grounded video feature associated with the current QA is treated as the positive sample $v^+$. This ultimately yields the following loss function:
\begin{equation}
    \mathcal{L}_{\textrm{Align}} = \lambda_1*\mathcal{L}_{\textrm{InfoNCE}}(v, l^+, l^-) + \lambda_2 * \mathcal{L}_{\textrm{InfoNCE}}(l_g, v^+, v^-)
\end{equation}
Besides, the loss function $\mathcal{L}_{\textrm{InfoNCE}}$ can be formulated as follows:
\begin{equation}
    \mathcal{L}_{\textrm{InfoNCE}}=-log(\frac{e^{sim(q, k^+)/\tau}}{e^{sim(q, k^+)/\tau}+\sum^N_{i=1}e^{sim(q, k_i^-)/\tau}})
\end{equation}
where the operator $sim$ is dot product.

\subsection{Explicit Causal Intervention Module}
After establishing the basic framework for VideoQG, the model can be formulated as:
\begin{equation}
    P(a|V,L)
    \label{eq:original_p}
\end{equation}
However, suffering from $Z$, the causal relations $V \to a$ and $L \to a$ may be ignored, leading to the spurious correlations $V \gets Z \to L$ and $Z \to a$, as shown in Figure~\ref{fig:method}.

Thus, Eq.~\ref{eq:original_p} is rewritten as:
\begin{equation}
    P(a|V,L,Z=z)P(Z=z|V,L)
    \label{eq:confounded_p}
\end{equation}
To alleviate this issue, causal intervention can be implemented by introducing the do calculus $do(\cdot)$.
By leveraging observable confounders in linguistics $Z_l$, such as the entities ``baby" and ``woman" and their semantic relations, back-door intervention~\cite{liu2022show, liu2023cross} can be implemented to block the path $Z_l \to L$ and mitigate confounding in $L$, as shown in Figure~\ref{fig:method} (b). The deconfounded probability of $P(a|V, do(L))$ can be formulated as:
\begin{equation}\small
\begin{split}
&P(a|V, do(L)) = \\
    &\sum_{Z_l=z_l}P(a|V, L, Z_l=z_l,Z_v=z_v)P(Z_l=z_l)P(Z_v=z_v|V)
    \label{eq:do_back}
\end{split}
\end{equation}
where $Z_l$ can be estimated from the clusters of the semantic structure graph $\widetilde{L}$, constructed by $L$. 
Specifically, we categorize the entities in Question $Q$ into three types: $sub$, $verb$, and $obj$, corresponding to the subject, verb, and object in the sentence. Based on these three elements, along with the $Q$ itself and its type, we construct a semantic structure graph. 
Additionally, because some datasets generate questions using a limited number of templates~\cite{wu2021star}, supplementary information from the answer $A$ is often necessary. 
Since the subject mentioned in the question is typically absent in the answer, we extract the verb and object from the answer and combine this information with the question to construct a more complete semantic structure graph, as shown in Figure~\ref{fig:module} (b). 

However, the visual confounders $Z_v$ still impede faithful answers, as shown in Eq.~\ref{eq:do_back}. 
Given the challenge of deconstructing visual confounders, we treat $v_t$ as the mediator $M$ and implement front-door causal intervention~\cite{liu2023cross,chen2023cross}, where the mediator can be described to provide insights into the mechanism of causal intervention. Then, $P(a|do(V), do(L))$ can be presented as the following:
\begin{equation}
\begin{split}
    &P(a|do(V), do(L)) = \\
    &P(a|do(V), do(L), M=v_t)P(M=v_t|do(V), do(L))
\end{split}
\label{eq:front_1}
\end{equation}
where $M$ is introduced by $V$ and $L$, while $L$ is deconfounded and there is no back-door path between $V$ and $M$. Thus, Eq.~\ref{eq:front_1} can be reformulated as:
\begin{equation}
    \sum_{v_t}P(a|do(M=v_t), do(L))P(M=v_t|V, do(L))
    \label{eq:front_2}
\end{equation}
which the probability of $P(a|do(M=v_t), do(L))$ can be formulated via back-door intervention applied at $M \gets V \gets Z \to a$ as:
\begin{equation}
\begin{split}
    &\sum_{\hat{v}}P(a|do(M=v_t), do(L), V=\hat{v}))P(V=\hat{v}|do(M=v_t))\\
    &=\sum_{\hat{v}}P(V=\hat{v})P(V=\hat{v}, M=v_t),
\end{split}
\label{eq:front_3}
\end{equation}
where $\hat{v}$ is the feature selected from $V$ to represent the overall distribution of the dataset. Combining Eq.~\ref{eq:front_2} and Eq.~\ref{eq:front_3}, we can further calculate Eq.~\ref{eq:front_1} as:
\begin{equation}
\begin{split}
&P(a|do(V), do(L)=\\
    &\sum_{\hat{v}}P(V=\hat{v})P(a|V=\hat{v}, M=v_t, do(L)*\\
    &\sum_{v_t}P(M=v_t|V=v, do(L))
\end{split}
\label{eq:final}
\end{equation}
where $*$ is the dot product, and the $\hat{v}$ can be estimated from the clusters center $\widetilde{V}$ of the frame features, embedded by the CLIP model. 
Finally, we employ the Normalized Weighted Geometric Mean (NWGM)~\cite{xu2015show} to estimate Eq.~\ref{eq:final} as formulated following:
\begin{equation}
\begin{split}
P(a|do(V), do(L)) \approx \textrm{Softmax}(g(V, L, \theta({\widetilde{L}}), \theta({\widetilde{V}}))).
\end{split}
\label{eq:nwgm}
\end{equation}
where $g(\cdot)$ is the network obtaining a debiased and accurate response, as shown in Figure~\ref{fig:module} (c). Considering whether the estimated cross-modal attention, represented by $v_t$, is more effective for task completion than the mean-based video feature $\Bar{v}$, which captures overall information, our ECI simulates intervention operations to improve the model's causal consistency.
Finally, the training target can be formulated as follows:
\begin{equation}
\mathcal{L}=\mathcal{L}_{CE} + \mathcal{L}_{Align}
\end{equation}
where $\mathcal{L}_{CE}$ denotes the cross-entropy loss for VideoQA.

\section{Experiments and Analysis}
\subsection{Datasets, Metrics, and Baselines}

\noindent \textbf{Datasets.} \textbf{1) NextGQA}NextGQA~\cite{xiao2024can} is a benchmark for the weakly supervised VideoQG task and extends the NextQA~\cite{xiao2021next}. It includes two types of questions: Causal (``why/how"), Temporal (``before/when/after"), and excludes Descriptive (“what/who/where”) mostly pertain to global content (e.g., “what event?”) or answers can be found almost throughout the whole video (e.g., “where is?”).
The dataset contains annotations for 10,531 valid time segments corresponding to 8,911 QA pairs and 1,557 videos, as shown in Table.~\ref{tab:data_gqa}. Most segments are shorter than 15 seconds, with an average duration of 7 seconds, significantly less than the total video length of approximately 40 seconds. These segments occupy an average of only 20\% of the full video, and their distribution across the left, middle, and right positions of the video is even. 

\textbf{2) STAR~\cite{wu2024star}} is a situated video question reasoning dataset built with naturally dynamic, compositional, and logical real-world videos, which consists of 4,901 videos, 60,206 questions, and corresponding time segments, as shown in Table~\ref{tab:data_star}. The questions are generated programmatically based on situational hypergraphs. Situated reasoning also requires structured situation comprehension and logical reasoning, which is a challenging benchmark for VideoQG models. It features four types of questions: Interaction, Sequence, Prediction, and Feasibility. Video scenes in the dataset are decomposed into hypergraphs containing atomic entities and relationships, such as actions, objects, and interactions. 

\begin{table}[]\scriptsize
    \centering
    \begin{tabular}{ccccccc}\hline
          &  Vid. & Que. & Seg. & Seg. Dur.(s) & Vid. Dir.(s) & Ratio (S./V.)\\
          \hline
         Train & 3,860 & 34,132 & - & - & 44.9 & -\\
         Val & 567 & 3,358 & 3,931 & 7.3 & 42.2 & 0.2\\
         Test & 990 & 5,553 & 6,600 & 6.7 & 39.5 & 0.2\\
         \hline
         Total & 5,417 & 43,043 & 10,531 & - & - & - \\
         \hline
    \end{tabular}
    \caption{Statistics of NExT-GQA dataset.}
    \label{tab:data_gqa}
\end{table}

\textbf{3) Comparison}
NextGQA is developed based on NextQA, with its text content also derived from the latter. NExT-QA focuses primarily on causal and temporal reasoning, posing questions like ``why" and ``how" to explore the sequence and causes of events. In contrast, STAR emphasizes contextual reasoning, involving logical inference based on the context and relationships within the video. While NExT-QA employs multiple-choice and open-ended questions, STAR offers a broader range of question types that require various forms of logical reasoning, such as predicting future actions or assessing the feasibility of events based on the video context. 

Furthermore, STAR's questions and answers are generated through automated scripts following standard templates, whereas NextQA relies on human annotations. This suggests that the automatically generated questions and answers in STAR may introduce more systematic and subtler biases. Consequently, as discussed in the main text, our CRA model shows more significant improvements on the STAR dataset compared to its performance on the NextGQA dataset. Additionally, in NextGQA, there are instances where a single QA pair corresponds to multiple time intervals.

\begin{table}[]\scriptsize
    \centering
    \begin{tabular}{ccccccc} \hline
          &  Vid. & Que. & Seg. & Seg. Dur.(s) & Vid. Dir.(s) & Ratio (S./V.)\\
          \hline
         Train & 3,032 & 45,731 & 45,731 & 11.5 & 30.0 & 0.39\\
         Val & 914 & 7,098 & 7,098 & 11.9 & 30.0 & 0.40\\
         Test & 955 & 7,377 & 7,377 & 11.6 & 29.7 & 0.40\\
         \hline
         Total & 4,901 & 60,206 & 60,206 & - & - & - \\
         \hline
    \end{tabular}
    \caption{Statistics of STAR dataset.}
    \label{tab:data_star}
\end{table}

\noindent \textbf{Metric}. We use the metrics from NextGQA, including the accuracy of Grounded QA (Acc@GQA), the accuracy of Video QA (Acc@VQA),  Intersection over Prediction (IoP), and Intersection over Union (IoU). 
The mean IoP (mIoP) and mean IoU (mIoU) refer to the average IoP and IoU values across multiple videos or samples. IoU@0.3/0.5 and Iop@0.3/0.5 are specific IoP and IoU metrics calculated with thresholds of 0.3 and 0.5, respectively.


\noindent \textbf{Baselines}. 
In our experiments, we select several high-performing visual language models as baselines, each representing different model architectures, language encoders, and visual encoders. These baseline models include: 
\begin{itemize}
    \item IGV~\cite{li2022invariant} and Sevila~\cite{yu2023self}: Originally designed for VQA tasks, we have modified these models to incorporate keyframe localization and enable accurate video localization of questions through post-hoc processing.
    \item VIOLEv2~\cite{fu2023empirical}: It uses a Swin Transformer for video encoding and BERT for text encoding. It then interacts with video and text features via a multimodal Transformer, showcasing superior video-text learning capabilities.
    \item VGT~\cite{xiao2022video}: It employs a graph transformer to capture visual objects while facilitating the comparison of relevance between video and text using a dual structure.
    \item Temp[Swin], Temp[CLIP], Temp[BLIP]~\cite{xiao2024can}: These models all feature a dual structure, differing only in the visual coders they use. ``PH" and ``NG+" represent different methods used to generate grounding intervals.
    \item FrozenBiLM~\cite{yang2022frozenbilm}: Utilizing frozen bidirectional language models to extract textual features, this model excels in VideoQA tasks, effectively combining text comprehension with video analysis.
    \item TimeCraft~\cite{liu2025timecraft}: a bi-directional reasoning framework for VideoQG, utilizing LLMs to expand the dataset and enable self-supervised temporal grounding and answering.
\end{itemize}
The selection and comparison of these baselines are crucial for evaluating the performance of various approaches and providing valuable insights for further optimization.

\subsection{Implementation Settings}
Following the setting of Temp[CLIP]~\cite{xiao2024can}, we utilized 32 evenly sampled frames from original videos. During the feature extraction, the CLIP-L model was frozen, while the Roberta model was finetuned. In the ECI module, we employed 512 visual clustering centers, and the QA part was similarly clustered to obtain 512 graph features in the LCI module. When calculating the align loss, we sampled 32 negative samples from the same batch. Additionally, the parameters $\lambda_1$ and $\lambda_2$ were set to 1 and 0.5, respectively. Other parameters were referred to the NextGQA benchmark and our model was trained with one RTX A800 GPU.
  
\subsection{Quantitative Analysis}
We divided the evaluation metrics into two categories: assessment of \textbf{Faithful Answer} (Acc@GQA) and evaluation of \textbf{Temporal Grounding} (IoP and IoU). 

\begin{table*}[]\scriptsize
    \centering
    \begin{tabular}{ccccccccccccc}
         \hline
         Method & Model & Vision & Text & Param. & Acc@GQA & Acc@QA & mIoP & TIoP@0.3 & TIoP@0.5 & mIoU & TIoU@0.3 & TIoU@0.5 \\
         \hline
         \rowcolor{gray!30}
         Human & - & - & - & - & 82.1  & 93.3 & 72.1 & 91.7 & 86.2 & 61.2 & 86.9 & 70.3 \\
         \rowcolor{gray!30}
         Random & - & - & - & - & 1.7 & 20.0 & 21.1 & 20.6 & 8.7 & 21.1 & 20.6 & 8.7 \\
         \hline
         \rowcolor{gray!30}
         \textit{SeViLA*} & BLIP-2 & \textit{ViT-G} & \textit{FT5} & \textit{4.1B} & \textit{16.6} & \textit{68.1} & \textit{29.5} & \textit{34.7} & \textit{22.9} & \textit{21.7} & \textit{29.2} & \textit{13.8} \\
         IGV  & - & ResNet & BT & - & 10.2 & 50.1 & 21.4 & 26.9 & 18.9 & 14.0 & 19.8 & 9.6\\
         MIST & Temp[CLIP] & BT & ViT-L & - & 12.1 & - & 23.6 & 29.3 & 20.7 & 11.4 & 16.3 & 7.0\\
         \hline
         PH & VIOLETv2 & VSWT & BT & - & 12.8 & 52.9 & 23.6 & 25.1 & 23.3 & 3.1 & 4.3 & 1.3 \\
         PH & VGT  & RCNN & RBT & - & 14.4 & 55.7  & 25.3 & 26.4 & 25.3 & 3.0 & 3.6 & 1.7 \\
         PH & Temp[Swin] & SWT & RBT & -  &  13.5 & 55.9  & 23.1 & 24.7 & 23.0 & 4.9 & 6.6 & 2.3 \\
         PH & Temp[CLIP]& Vit-B & RBT & -  & 14.7 & 57.9  & 24.1 & 26.2 & 24.1 & 6.1 & 8.3 & 3.7 \\
         PH & Temp[BLIP]& ViT-B & RBT & - & 14.9 & 58.5  & 25.0 & 27.8 & 25.3 & 6.9 & 10.0 & 4.5\\
         \hline
         PH & Temp[CLIP]& ViT-L & RBT & 130.3M & 15.2 & 59.4 & 25.4 & 28.2 & 25.5 & 6.6 & 9.3 & 4.1 \\
         NG+ & Temp[CLIP] & ViT-L & RBT & 130.6M & \underline{16.0} & 60.2 & 25.7 & 31.4 & 25.5 & 12.1 & 17.5 & 8.9 \\
         \rowcolor{red!10}
         TimeCraft  & Temp[CLIP] & ViT-L & RBT & 130.6M & \textbf{18.2} & - & \underline{28.1} & \textbf{35.1}& \underline{27.8} & \textbf{15.6}& \underline{21.2} & \underline{9.6}\\
         \rowcolor{green!10}
         CRA (ours) & Temp[CLIP] & ViT-L & RBT & 145.1M & \textbf{18.2} & \textbf{61.1}  & \textbf{28.6} & \underline{34.3} & \textbf{28.5} & \underline{14.2} & \textbf{21.4} & \textbf{10.6}\\
         \hline
         PH & FrozenBiLM& ViT-L & DBT & 1.2B & 15.8 & 69.1 & 22.7 & 25.8 & 22.1 & 7.1 & 10.0 & 4.4\\
         NG+ & FrozenBiLM & ViT-L & DBT & 1.2B & 17.5 & \textbf{70.8} & 24.2 & 28.5 & 23.7 & 9.6 & 13.5 & 6.1\\
         \rowcolor{red!10}
         TimeCraft & FrozenBiLM & ViT-L & DBT & 1.2B & \underline{18.5} & -  & \underline{26.3} & \textbf{32.7}& \underline{24.9}& \underline{13.2} & \underline{18.6} & \underline{8.4}\\
         \rowcolor{green!10}
         CRA (ours) & FrozenBiLM & ViT-L & DBT & 1.2B & \textbf{18.8} & \underline{70.2} & \textbf{26.5} & \underline{32.6} & \textbf{25.9} & \textbf{13.5} & \textbf{20.1} & \textbf{9.6}\\
         \hline
    \end{tabular}
        \vspace{-8pt}
    \caption{VideoQG performance on NextGQA test set. BT: BERT. RBT: RoBERTa. DBT: DeBERTa-V2-XL. FT5: Flan-T5-XL. Random: always choose the same answer ID and return the whole video duration as the grounding result. *: pre-train on video-language grounding dataset. It is important to note that the data with gray shading in the table is provided for reference only and is not included in the comparisons. Both IGV and SeViLA incorporate explicit temporal grounding in their training. \colorbox{red!10}{TimeCraft} introduces extra data constructed by Llama-2 13B. We compared the performance of methods using Temp[CLIP] and FrozenBiLM as backbones, respectively.}
    \label{tab:main_result_gqa}
\end{table*}

\begin{table}[]\scriptsize
    \centering
    \setlength{\tabcolsep}{4pt}
    \begin{tabular}{lcccc}         \hline
         Method         & Acc@GQA &  Acc@VQA & TIoP@0.5 & TIoU@0.5 \\
         \hline
         \rowcolor{gray!30}
         IGV                      & 13.1  & 31.7 & 42.8 & 1.7 \\
         \rowcolor{gray!30}
         SeViLA*                  & 17.0  & 45.5 & 37.6 & 19.5\\
         \hline
         Temp[CLIP] (NG+)         & 24.4  & 57.3            & 41.4 & 4.7 \\
         \rowcolor{green!10}
        \textbf{Temp[CLIP] (CRA)} & \textbf{26.8} &           \textbf{58.6}     & \textbf{44.5} & \textbf{5.5} \\
         \hline
         FrozenBiLM (NG+)         & 25.8  &      60.1 & 40.9 & \textbf{7.8} \\
         \rowcolor{green!10}
         \textbf{FrozenBiLM (CRA)} & \textbf{27.5} &           \textbf{60.5}     & \textbf{43.1} & 5.1 \\
         \hline
    \end{tabular}
        \vspace{-8pt}
    \caption{VideoQG performance on STAR val set. }
    \label{tab:main_result_star}
\end{table}

\subsubsection{Faithful Answer.}
As shown in Table~\ref{tab:main_result_gqa}, 
our method outperforms the comparison methods, particularly the Temp[CLIP] (NG+), where we achieve significant improvements despite having a comparable model size (only larger by 15M parameters). Specifically, the proposed method is 3\% higher in Acc@GQA compared to Temp[CLIP] (NG+). Besides, when using FrozenBiLM as the backbone, the proposed method still surpasses the current SOTA model, TimeCraft, by 0.3\%. This demonstrates that the proposed method not only provides higher accuracy in VideoQA tasks but also proves more effective in temporal grounding, leading to overall superior performance. 

Additionally, we observed that in the experiments with FrozenBiLM as the backbone, the Acc@VQA performance consistently outperforms Temp[CLIP], while the gap in Acc@GQA is not as large. This suggests that while larger models excel in accuracy for question answering, they struggle with faithful temporal grounding, resulting in a higher incidence of unfaithful answers. This advantage stems from the model’s pre-training on larger datasets, although this pre-training does not account for causal consistency. A similar trend was observed in the SeViLA model, which achieved an Acc@VQA of 68.1\% but only an Acc@GQA of 16.6\%. Moreover, TimeCraft utilizes Llama-2 13B to generate QA pairs for specific events in the video and match these with the original QA in the dataset through a Dual-QA approach, using a bidirectional generation method to align text-image information. Despite this, our approach is rooted in extracting the causal structure inherent in the dataset itself, going beyond the LLM-based prior of TimeCraft, regardless of whether Temp[CLIP] or FrozenBiLM is used as the backbone.

Furthermore, for the Random method, we observe that its IoU is not inferior to that of explicitly trained models like IGV. However, its Acc@GQA metrics are significantly lower than all the methods in the table, highlighting the necessity for models to ensure that their answers are not only correct but also grounded in visual cues consistent with causal reasoning. Our CRA effectively prevents the model from relying on spurious correlations, instead fostering a deeper and more reliable connection between the video content and the generated answers. This assertion is further validated by results on the STAR dataset, where our CRA also achieves the best Acc@GQA score, as shown in Table~\ref{tab:main_result_star}. 

Additionally, we employ metrics similar to the Acc@GQA metric. Samples with an IoP $<0.3$ can be noted as follows: incorrect answers are denoted bias errors; correct answers are considered unfaithful answers. We effectively reduce bias-induced errors and decrease the occurrence of unfaithful answers, aligning with the improvements observed in the Acc@GQA metric (Table~\ref{tab:de_bias}).

\begin{table}[]\scriptsize
    \centering
    \begin{tabular}{ccccc}
        \hline
        Method & Acc@GQA$\uparrow$ & Acc@QA$\uparrow$ & Bias Error$\downarrow$ & Unfaithful$\downarrow$\\
        \hline
        PH  & 15.3 & 59.0 & 28.5 & 41.4 \\
        CRA & 18.2(+2.9)& 61.1(+2.1) & 27.4(-1.1) & 40.0(-1.4)\\
        \hline
    \end{tabular}
    \vspace{-10pt}
    \caption{Quantify the de-bias}
    \vspace{-10pt}
    \label{tab:de_bias}
\end{table}

\subsubsection{Temporal Grounding.}

Table~\ref{tab:main_result_gqa} and Table~\ref{tab:main_result_star} reveal that TimeCraft using LLMs to construct pseudo-labels for cross-modal alignment, our approach achieves superior IoP and IoU metrics, especially with the more stringent IoP@0.5 and IoU@0.5. This indicates that our CRA more efficiently achieves multimodal alignment, mitigating multimodal bias while significantly enhancing causal consistency for VideoQG.
Additionally, we observed that the Naive Gauss (NG) strategy employed by Temp[CLIP], while simple and effective, generates grounding intervals centered around keyframes. However, keyframes related to the question are not always located at the center of the interval, they may be positioned towards either end of the interval. For instance, in the interval from Figure~\ref{fig:intro_causal}, dominant information (e.g., the interaction between woman and baby) appears in the first frame.

\begin{figure}[!t]
    \centering
    \includegraphics[width=1\linewidth]{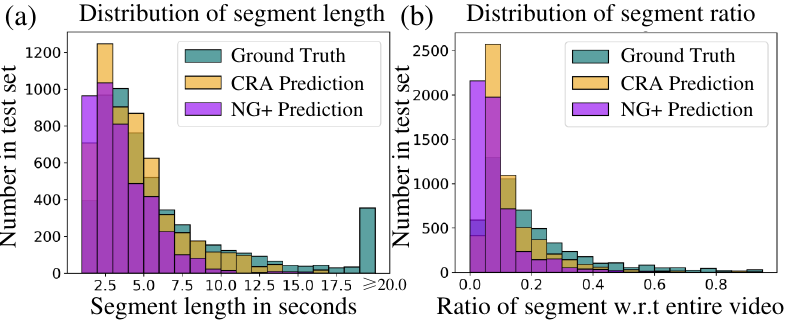}
        \vspace{-15pt}
    \caption{(a) shows the distribution of segment length of CRA, Temp[CLIP] (NG+), and Ground Truth on NextGQA dataset. (b) shows the distribution of segment ratio of CRA, Temp[CLIP] (NG+), and Ground Truth. The hierarchical bin can be compared intuitively.}
    \vspace{-10pt}
    \label{fig:analysis}
\end{figure}

\begin{figure*}[!t]
    \centering
    \includegraphics[width=1\linewidth]{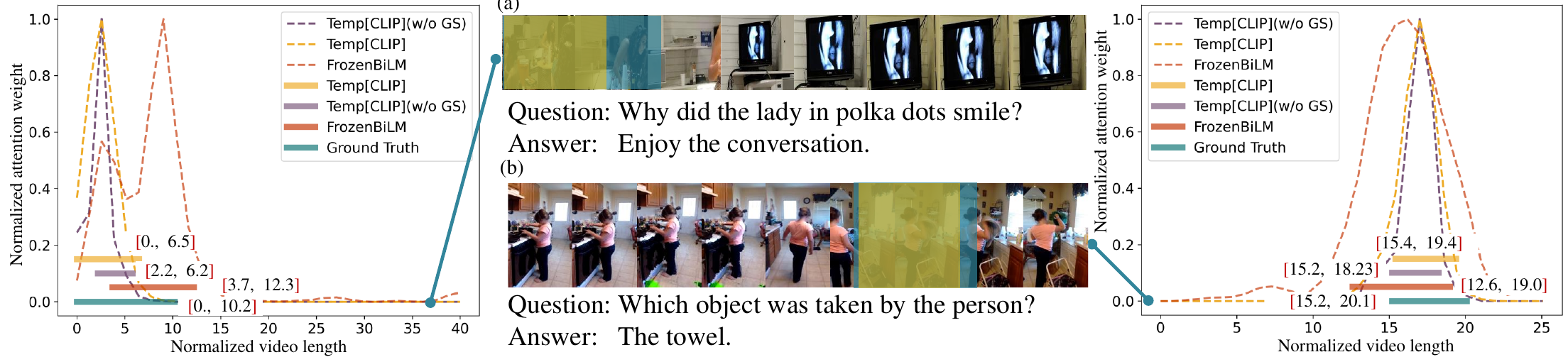}
    \vspace{-15pt}
    \caption{Visualization examples in NextGQA dataset (a) and STAR dataset (b). The numbers ([start time, end time]) indicate the interval.}
    \label{fig:vis_result}
\end{figure*}

To validate this, we compare the accuracy of time intervals generated by four different strategies, as shown in Table~\ref{tab:time_gen_policy}. The PH strategy still yields good performance; however, its IoU@0.5 metric is the worst. Moreover, compared to GSG w/o GS, the latter achieves higher Acc@GQA with lower Acc@VQA, indicating that grounding solely derived from video saliency is unreliable. This is because a single video may contain multiple events, each corresponding to different QA pairs. Additionally, we compared GSG w/o GS$\dagger$, where removing Gaussian smoothing resulted in the model's inability to refine accurate visual features based on coarse attention. Although its IoU@0.5 is consistent with that of GSG w/o GS, its overall performance is the worst. Notably, when Gaussian smoothing was applied, there was a significant improvement in IoU@0.5. This suggests that the Gaussian smoothing in GSG plays a critical role in suppressing time-irrelevant noise and enhancing the model’s focus on key regions.

\begin{table}[]\scriptsize
    \centering
    \setlength{\tabcolsep}{4pt}
    \begin{tabular}{lcccc}         \hline
         Method & Acc@GQA & Acc@VQA & TIoP@0.5 & TIoU@0.5\\
         \hline
         CRA (PH) & 16.4  & 60.1 & 26.7 & 8.0\\
         CRA (GSG w/o GS) & 16.6 & 59.8 & 27.3 & 8.4\\ 
         CRA (GSG w/o GS$\dagger$) & 15.8& 58.2 & 25.6 & 8.4\\
         CRA (GSG) & \textbf{18.2}  & \textbf{61.1}& \textbf{28.5} & \textbf{10.6}\\
         \hline
    \end{tabular}
        \vspace{-8pt}
    \caption{Different time interval generation policies on NextGQA dataset. PH denotes the post-hoc analysis without the GSG module, and GSG w/o GS denotes our proposed temporal grounding approach, which omits Gaussian smoothing during training. SG w/o GS$\dagger$ indicates GSG with Gaussian smoothing during training but inference without it.}
    \vspace{-10pt}
    \label{tab:time_gen_policy}
\end{table}


\subsubsection{Detailed Analysis on NextGQA}

As shown in Table~\ref{tab:appendix_res_gqa}, a more detailed analysis was conducted based on different question types in the NextGQA benchmark. This benchmark includes two main categories of questions: causal questions, comprising 3,252 examples (58.6\% of the total), as mentioned in the dataset analysis section, and temporal questions, comprising 2,301 examples (41.4\% of the total). A comparison of these results with the overall performance indicates that the CRA framework achieves superior performance on causal questions. Notably, while Temp[CLIP] and FrozenBiLM achieve identical Acc@GQA scores, Temp[CLIP] exhibits significantly higher IoP@0.5, whereas FrozenBiLM outperforms Acc@VQA. This suggests that larger models, despite leveraging data priors learned from extensive datasets, also introduce more pronounced biases. Nevertheless, the CRA framework significantly mitigates these biases on the FrozenBiLM~\cite{yang2022frozenbilm} model when compared to the NG+ method~\cite{xiao2024can}. 

Additionally, for the temporal question category, CRA achieves the highest Acc@GQA scores across both models. This indicates that CRA demonstrates a higher degree of causal consistency between the retrieved video segments and the answers in the VideoQG task. Regarding the IoP@0.5 metric, the difference between Temp[CLIP] and FrozenBiLM is minimal, suggesting that temporal tasks are less affected by biases introduced during large-scale model pretraining. Consequently, CRA demonstrates robust improvements across various scenarios.

\begin{table*}[]\scriptsize
    \centering
    \begin{tabular}{c|cccccc}\hline
         Que.Type & Method & Model & Acc@GQA  & Acc@VQA  & IoP@0.5 & IoU@0.5 \\
         \hline
         \multirow{2}{*}{Causal}    & NG+ & Temp[CLIP] & 18.2 & 60.3 & 29.5 & 9.8 \\
                                    & NG+ & FrozenBiLM & 17.6 & 70.9 & 23.3 & 8.1 \\
         3,252                      & \cellcolor{green!10}CRA & \cellcolor{green!10}Temp[CLIP] & \cellcolor{green!10}\textbf{20.2} & \cellcolor{green!10}60.7 & \cellcolor{green!10}\textbf{31.9} & \cellcolor{green!10}\textbf{10.8}\\ 
         58.6\%                     & \cellcolor{green!10}CRA & \cellcolor{green!10}FrozenBiLM & \cellcolor{green!10}\textbf{20.2} & \cellcolor{green!10}\textbf{71.3} & \cellcolor{green!10}27.4 & \cellcolor{green!10}10.0 \\
         \hline
         \multirow{2}{*}{Temporal}  & NG+ & Temp[CLIP] & 12.7 & 60.5 & 20.9 & 8.4\\
                                    & NG+ & FrozenBiLM & 14.1 & 68.6 & 19.6 & 8.3 \\
         2,301                      & \cellcolor{green!10}CRA & \cellcolor{green!10}Temp[CLIP] & \cellcolor{green!10}15.4 & \cellcolor{green!10}61.8 & \cellcolor{green!10}23.7 & \cellcolor{green!10}\textbf{10.3}\\
         41.4\%                     & \cellcolor{green!10}CRA & \cellcolor{green!10}FrozenBiLM & \cellcolor{green!10}\textbf{16.8} & \cellcolor{green!10}\textbf{68.9} & \cellcolor{green!10}\textbf{23.8} & \cellcolor{green!10}9.0 \\
         \hline
         \multirow{4}{*}{Total}     & NG+ & Temp[CLIP] & 15.9 & 60.2 & 25.9 & 9.2\\
                                    & NG+ & FrozenBiLM & 16.1 & 69.9 & 21.8 & 8.2 \\
                                    & \cellcolor{green!10}CRA & \cellcolor{green!10}Temp[CLIP] & \cellcolor{green!10}18.2 & \cellcolor{green!10}61.1  & \cellcolor{green!10}\textbf{28.5} & \cellcolor{green!10}\textbf{10.6} \\
                                    & \cellcolor{green!10}CRA & \cellcolor{green!10}FrozenBiLM & \cellcolor{green!10}\textbf{18.8} &\cellcolor{green!10} \textbf{70.3} & \cellcolor{green!10}25.9 & \cellcolor{green!10}9.6 \\
         \hline
    \end{tabular}
    \caption{Comparison with state-of-the-art methods on NextGQA test set. We train the Temp[CLIP](NG+) and FrozonBiLM(NG+) models on the NextGQA dataset via the official code.}
    \label{tab:appendix_res_gqa}
\end{table*}

\subsubsection{Detailed Analysis on STAR}

In our analysis of the STAR dataset, we categorize the questions into four types: Interaction (2,398 questions, accounting for 33.8\% of the total), Sequence (3,586 questions, 50.5\%), Prediction (624 questions, 8.8\%), and Feasibility (490 questions, 6.9\%), as shown in Table~\ref{tab:appendix_res_star}. It is evident that Interaction and Sequence questions dominate the dataset, comprising over 85\% of the questions and significantly influencing the overall performance.

Firstly, for Interaction-type questions, although our approach demonstrates average performance in the Acc@VQA and Acc@GQA metrics, it achieves the best results in the temporal grounding task. This indicates that our model can more accurately locate relevant information when interpreting interactions between people and objects in the video. However, this strong grounding performance does not translate into causal consistency in the answers.

Sequence-type questions stand out, although the proposed method achieves an Acc@VQA score 1.6\% lower than FrozenBiLM (NG+) on the Temp[CLIP] model, it surpasses the latter by 6.7\% in IoP@0.5 and 2.7\% in Acc@GQA. These results highlight the model's exceptional performance in handling temporal reasoning tasks, demonstrating a superior ability to capture the sequence and logic of events. This leads to a deeper understanding of video content and a high degree of causal consistency.

For predictive questions, overall performance is slightly better than that for sequential questions. As shown in the table, the performance of FrozenBiLM consistently surpasses that of Temp[CLIP], including in the IoP@0.5 metric. This suggests that larger-scale models exhibit stronger reasoning capabilities for predicting future events, a benefit derived from the prior knowledge embedded in their training data.

\begin{table*}[]\scriptsize
    \centering
    \begin{tabular}{c|cccccc}\hline
         Que.Type & Method & Model & Acc@GQA  & Acc@VQA & IoP@0.5 & IoU@0.5 \\
         \hline
         Interaction            & NG+ & Temp[CLIP] & 12.5 & 52.3 & 23.6 & 5.6 \\
         2,398                  & NG+ & FrozenBiLM & 13.4 & 54.3 & 22.8 & 7.3 \\
         33.8\%                 &\cellcolor{green!10}CRA & \cellcolor{green!10}Temp[CLIP] & \cellcolor{green!10}14.1 & \cellcolor{green!10}53.3 & \cellcolor{green!10}\textbf{25.5} & \cellcolor{green!10}\textbf{7.5} \\
                                &\cellcolor{green!10}CRA & \cellcolor{green!10}FrozenBiLM & \cellcolor{green!10}\textbf{14.8} & \cellcolor{green!10}\textbf{54.7} & \cellcolor{green!10}25.4 & \cellcolor{green!10}5.6 \\
         \hline
         Sequence               & NG+ & Temp[CLIP] & 32.0 & 59.5 & 53.2 & 4.2 \\
         3,586                  & NG+ & FrozenBiLM & 32.2 & 62.5 & 50.6 & \textbf{8.2} \\
         50.5\%                 & \cellcolor{green!10}CRA & \cellcolor{green!10}Temp[CLIP] & \cellcolor{green!10}\textbf{34.9} & \cellcolor{green!10}60.9 & \cellcolor{green!10}\textbf{57.3} & \cellcolor{green!10}3.8 \\
                                & \cellcolor{green!10}CRA & \cellcolor{green!10}FrozenBiLM & \cellcolor{green!10}34.0 & \cellcolor{green!10}\textbf{62.9} & \cellcolor{green!10}52.1 & \cellcolor{green!10}4.0 \\
         \hline
         Prediction             & NG+ & Temp[CLIP] & 33.2 & 61.5 & 51.9 & 5.3 \\
         624                    & NG+ & FrozenBiLM & 37.5 & 64.8 & 57.9 & \textbf{10.4}\\
         8.8\%                  & \cellcolor{green!10}CRA & \cellcolor{green!10}Temp[CLIP] & \cellcolor{green!10}36.5 & \cellcolor{green!10}62.8 & \cellcolor{green!10}56.9 & \cellcolor{green!10}4.6 \\
                                & \cellcolor{green!10}CRA & \cellcolor{green!10}FrozenBiLM & \cellcolor{green!10}\textbf{40.9} & \cellcolor{green!10}\textbf{65.9} & \cellcolor{green!10}\textbf{60.6} & \cellcolor{green!10}7.2 \\
         \hline
         Feasibility            & NG+ & Temp[CLIP] & 16.1 & 61.5 & 28.4 & 4.2 \\
         490                    & NG+ & FrozenBiLM & 21.6 & \textbf{66.1} & 33.9 & 7.1 \\
         6.9\%                  & \cellcolor{green!10}CRA & \cellcolor{green!10}Temp[CLIP] & \cellcolor{green!10}17.6 & \cellcolor{green!10}63.3 & \cellcolor{green!10}27.8 & \cellcolor{green!10}\textbf{8.6} \\
                                & \cellcolor{green!10}CRA & \cellcolor{green!10}FrozenBiLM & \cellcolor{green!10}\textbf{25.3} & \cellcolor{green!10}64.1 & \cellcolor{green!10}\textbf{41.8} & \cellcolor{green!10}7.6 \\
         \hline
         \multirow{4}{*}{Total} & NG+ & Temp[CLIP] & 24.4 &  57.3 & 41.4 & 4.7 \\
                                & NG+ & FrozenBiLM & 25.8 &  60.1 & 40.9 & \textbf{7.8} \\
                                & \cellcolor{green!10}CRA & \cellcolor{green!10}Temp[CLIP] & \cellcolor{green!10}26.8 &  \cellcolor{green!10}58.6 & \cellcolor{green!10}\textbf{44.5} & \cellcolor{green!10}5.5 \\
                                & \cellcolor{green!10}CRA & \cellcolor{green!10}FrozenBiLM & \cellcolor{green!10}\textbf{27.5} &  \cellcolor{green!10}\textbf{60.5} & \cellcolor{green!10}43.1 & \cellcolor{green!10}5.1 \\
         \hline
    \end{tabular}
    \caption{Comparison with state-of-the-art methods on STAR val set because the Acc.@GQA metric can not be calculated on the private test set.}
    \label{tab:appendix_res_star}
\end{table*}

Furthermore, this perspective is further validated in feasibility-related questions. Such questions are characterized by their diversity and complexity, involving not only directly observable information from videos but also implicit conditions and assumptions. These questions typically require a deep understanding of the video context and the ability to infer whether a given scenario is plausible in real life. This often demands sophisticated logical reasoning and consideration of multiple factors. For instance, a question might require the model to determine the feasibility of an action under specific conditions, necessitating not only an understanding of the video content but also reasoning about underlying physical principles and common-sense knowledge.
The inherent difficulty of these questions explains why the large-scale FrozenBiLM model performs best in this category. Notably, with the enhancement of the CRA framework, FrozenBiLM achieves an impressive IoP@0.5 score of 41.8\%. This finding motivates further development of the CRA framework with even larger models to enhance its capability to handle such complex reasoning tasks.

\subsection{Qualitative Analysis}
Despite most time intervals in the ground truth being relatively short, a significant number exceeds the 20s, as shown in Figure~\ref{fig:analysis} (a). Our method and NG+ tend to generate time intervals concentrated around a short duration of approximately 2.5s, neglecting the estimation of longer intervals. Further analysis reveals that in the ground truth, intervals shorter than 2.5s are fewer than those between 2.5s and 5s, while the estimated intervals are predominantly clustered around 2.5s. This is because the estimated attention peaks are generally narrow, and inaccurate attention can lead to imprecise short intervals. However, as shown in Figure~\ref{fig:analysis} (b), NG+, which does not incorporate causal intervention, performs noticeably worse than CRA. This indicates that the attention weights in NG+ are overly focused, while our method with causal intervention is more robust. Our approach more effectively observes information around keyframes, better captures relevant details, and aligns more closely with the true distribution of time intervals.

\subsubsection{Visualization of CRA on NextGQA dataset}
\begin{figure*}[!t]
    \centering
    \includegraphics[width=1\linewidth]{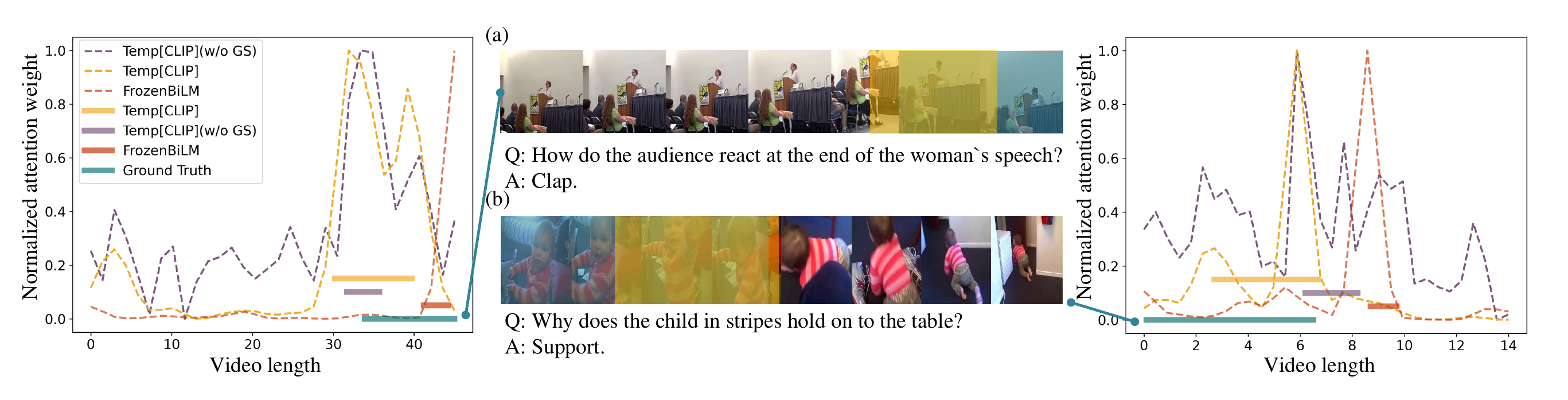}
    \vspace{-20pt}
    \caption{Visualization examples in NextGQA dataset. The numbers ([start time, end time]) indicate the interval range.}
    \label{fig:vis_result_next}
\end{figure*}

\begin{figure*}[]
    \centering
    \includegraphics[width=1\linewidth]{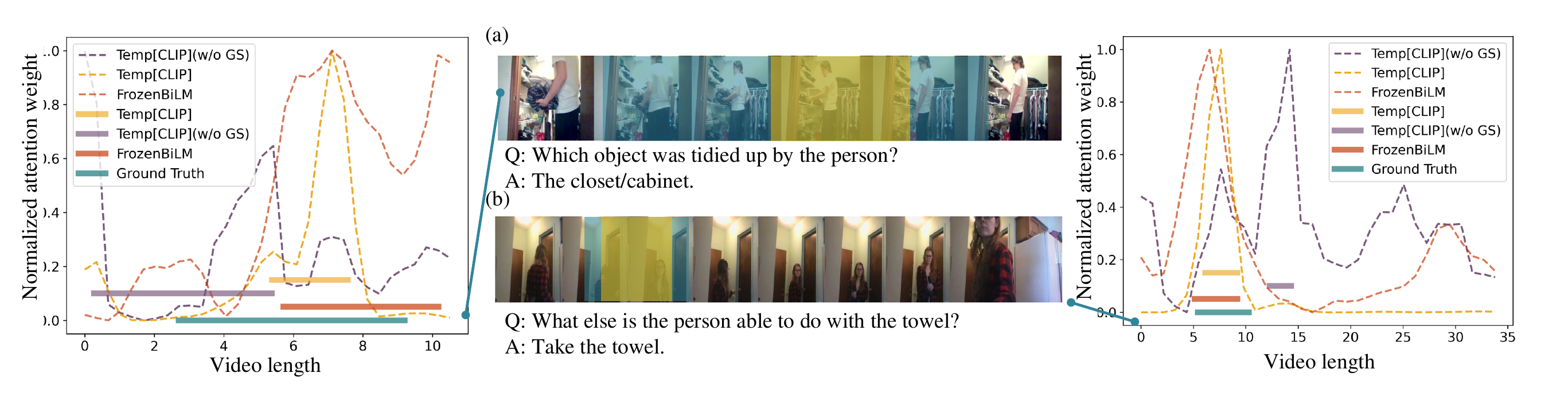}
    \vspace{-20pt}
    \caption{Visualization examples in STAR dataset. The numbers ([start time, end time]) indicate the interval range.}
    \label{fig:vis_result_star}
\end{figure*}

As shown in Figure \ref{fig:vis_result_next}, we present the visualization results of CRA on the NextGQA test set. In Figure \ref{fig:vis_result_next}(a), the question belongs to the Temporal category. After removing Gaussian smoothing, the attention weights exhibit significant oscillations along the temporal axis, preventing the model from effectively estimating the intervals. However, the proposed GSG module successfully mitigates these noise effects, enabling the accurate localization of relevant intervals, thereby improving both IoP@0.5 and IoU@0.5 performance. Nevertheless, the Temp[CLIP] model, despite these enhanced weights, still fails to provide a correct answer to this question. In contrast, the FrozenBiLM model delivers the correct answer by relying solely on the final frame. From the attention weights, it is evident that the model is highly confident, producing a single narrow peak and identifying a short temporal interval. This behavior highlights the inherent bias of large-scale models.

Similarly, as shown in Figure \ref{fig:vis_result_next}(b), our method effectively identifies relevant intervals and correctly answers causal questions. However, it is notable that while FrozenBiLM also answers correctly, it confidently attends to incorrect visual information. This further confirms the more severe spurious correlations introduced by data biases in large models. Additionally, comparing the provided ground truth reveals that our method is not entirely incorrect. The video segment estimated by CRA is sufficient to support the answer, while the ground truth interval appears unnecessarily redundant. This observation underscores the greater importance of IoP@0.5 compared to IoU@0.5, as the task prioritizes the precision of interval estimation.

\subsubsection{Visualization of CRA on STAR dataset}
As illustrated in Figure~\ref{fig:vis_result_star}, we present the visualization of CRA on the STAR dataset. As mentioned earlier, Figure~\ref{fig:vis_result_star} (a) depicts a scenario where a man is organizing a wardrobe, an activity that spans the entire video. The ground truth segment is centered within the video, which is a reasonable choice. However, the video segments adopted by various methods appear sufficient to serve as the basis for answering the question. This suggests that the dataset may contain some annotation noise and that the evaluation methods could have certain limitations. On the other hand, in the example of a Feasibility-category question as shown in Figure~\ref{fig:vis_result_star} (b), the effectiveness of Gaussian smoothing is reaffirmed. This approach effectively suppresses noise and facilitates better multi-modal alignment.

\subsection{Ablation Studies}
As shown in Table~\ref{tab:ablation_result_gqa}, our CRA framework leverages cross-modal attention in the GSG module to effectively retrieve intervals most relevant to the current QA, rather than relying solely on visual saliency. When the GSG module is removed, the IoU@0.5 performance significantly drops (10.6 $\to$ 8.0). 
This decline is attributed to the fact that the GSG module integrates cross-modal information and uses Gaussian smoothing for denoising, enabling broader and more robust attention, which results in more accurate time intervals, as illustrated in Figure~\ref{fig:vis_result}. Furthermore, although the attention and intervals do not show significant differences before and after Gaussian smoothing is applied, its impact on the subsequent results is substantial, as shown in Table~\ref{tab:time_gen_policy}.
Notably, although the interval generated by FrozenBiLM in the figure overlaps more with the ground truth area, its attention distribution is clearly less focused, with two distinct peaks. This suggests that while larger models may capture a wider and finer-grained range of information, not all of this information is beneficial for grounding.

\begin{table}[!t]\scriptsize
    \centering
    \setlength{\tabcolsep}{8pt}
    \begin{tabular}{lcccc}         \hline
         Method & Acc@GQA & Acc@VQA & TIoP@0.5 & TIoU@0.5 \\
         \hline
         CRA        & 18.2                   & 61.1                   & 28.5                   & 10.6\\
         w/o GSG    & 16.4 ($\downarrow$1.8) & 60.1 ($\downarrow$1.0) & 26.7 ($\downarrow$1.8) & 8.0  ($\downarrow$2.6)\\
         w/o CMA    & 16.2 ($\downarrow$2.0) & 60.0 ($\downarrow$1.1) & 26.3 ($\downarrow$2.2) & 9.3  ($\downarrow$1.3)\\
         w/o Causal & 17.0 ($\downarrow$1.2) & 60.1 ($\downarrow$1.0) & 27.6 ($\downarrow$0.9) & 9.5  ($\downarrow$1.1)\\
         w/o LCI    & 17.8 ($\downarrow$0.4) & 60.9 ($\downarrow$0.2) & 28.2 ($\downarrow$0.3) & 10.4 ($\downarrow$0.2)\\
         w/o ECI    & 17.3 ($\downarrow$0.9) & 60.5 ($\downarrow$0.6) & 27.3 ($\downarrow$1.2) & 9.6  ($\downarrow$1.0)\\ 
         \hline
    \end{tabular}
    \caption{Ablation studies of CRA on NextGQA dataset.}
    \label{tab:ablation_result_gqa}    
\end{table}

Moreover, removing the CMA module results in a significant loss in both Acc@GQA and IoP@0.5, with decreases of 2\% and 2.22\%, respectively. This highlights the critical role of the CMA module in the VideoQG task, demonstrating that CMA effectively aligns multimodal information through bidirectional contrastive learning, enhancing the model’s ability to infer correct answers from video content without relying on extensive labeled data. Therefore, the CMA module is indispensable for achieving stability and accuracy in weakly-supervised VideoQG tasks.

The removal of the causal module also leads to a noticeable decline in overall metrics. However, unlike the CMA module, the loss in IoP@0.5 is smaller, while Acc@GQA suffers a larger drop. This result aligns with the intention of using the ECI module to intervene in cross-modal features, aiming to establish more fundamental causal relations between these features and the answers, thereby enhancing causal consistency between VideoQA and Video Temporal Grounding. Furthermore, we conduct additional ablation experiments on the causal modules for both the language and visual modalities, as shown in Table~\ref{tab:ablation_result_gqa}. The removal of visual causal intervention results in a more severe performance drop (1.2\%), particularly in the IoP@0.5 metric. This indicates that the explicit causal intervention for visual modality efficiently aligns causal relations, and its effectiveness is clearly reflected in the IoP@0.5 metric.
\vspace{-1pt}

\subsubsection{Ablation of LLM}
Additionally, we conducted experiments using more advanced LLMs as the text model. Within the framework of FrozenBiLM, DeBERTa-V2-XL was replaced with Qwen2.5-1.5B, with no alterations made to the remaining components. As observed from Table~\ref{tab:llm}, our approach achieved a significant improvement in the Acc@GQA and IoP@0.5 metrics despite a marginal increase (0.2) in the Acc@QA metric. This indicates that our method effectively focuses on capturing cross-modal causality and enhances the causal consistency of the VideoQA task rather than relying on spurious correlations for superficial performance gains. Additionally, since FrozenBiLM employs weights fine-tuned from cross-modal pretraining, and our experiment utilized only the LLM version of Qwen2.5, a performance gap is expected.

\begin{table}[!t]\scriptsize
\setlength{\tabcolsep}{4pt}
    \centering
    \begin{tabular}{ccccc}
        \hline
        Method & Text & Acc@GQA & Acc@QA & TIoP@0.5\\
        \hline
        NG+ & Qwen2.5-1.5B & 15.0 & 65.2 & 21.5 \\
        CRA & Qwen2.5-1.5B & 16.5 & 65.4 & 23.1 \\
        \hline
    \end{tabular}
    \caption{Ablation of the LLM on NExT-GQA dataset}
    \label{tab:llm}
\end{table}

\subsubsection{Compare Same Backbone with IGV}
Most existing work implicitly performs deconfounding, with the effectiveness of their causal modules primarily evaluated using the Acc@VQA metric. Our CRA integrates causal front-door intervention with the Grounded VideoQA task, enabling the performance of the indirectly trained Temporal Grounding to directly quantify the effectiveness of the causal module.
Additionally, while IGV constructs a causal model based on scene invariance, our CRA achieves finer alignment through causal intervention compared to the coarse-grained segmentation and recombination of video clips. For a fair comparison, we employed the same backbone as IGV for experimentation, and our method still demonstrated superior performance, as shown in Table~\ref{tab:igv}. 

\begin{table}[]\scriptsize
\setlength{\tabcolsep}{4pt}
    \centering
    \begin{tabular}{cccccc}
        \hline
        Method & Vision & Text & Acc@GQA & Acc@QA & TIoP@0.5 \\
        \hline
        IGV & ResNet & BT & 10.2 & 50.1 & 18.9\\
        CRA & ResNet & BT  & 13.2 & 51.5 & 23.9\\
        \hline
    \end{tabular}
    \caption{Comparison with IGV on NExT-GQA dataset}
    \label{tab:igv}
\end{table}

\section{Conclusion}

This paper aims to perform cross-modal causal relation alignment to interpret grounded video segments during the question-answering process in VideoQG. We propose a weakly supervised VideoQG model that leverages existing VideoQA datasets and 
introduce cross-modal alignment to further enhance feature alignment across modalities. 
Additionally, 
we incorporate an explicit causal intervention module to eliminate spurious cross-modal correlations, thereby improving the causal consistency between question-answering and temporal grounding. 
Extensive experiments on NextGQA and STAR datasets demonstrate the effectiveness of our approach. The promising results, including high Acc@GQA and IoU@0.5 scores, show that our CRA achieves robust and reliable VideoQG performance, effectively grounding visual content and supporting accurate question reasoning.

{
    \small
    \bibliographystyle{ieeenat_fullname}
    \bibliography{main}
}


\end{document}